# Random Forest Ensemble of Support Vector Regression Models for Solar Power Forecasting

Mohamed Abuella, Student Member, IEEE
Energy Production and Infrastructure Center
Department of Electrical and Computer Engineering
University of North Carolina at Charlotte
Charlotte, USA
Email: mabuella@uncc.edu

Badrul Chowdhury, Senior Member, IEEE
Energy Production and Infrastructure Center
Department of Electrical and Computer Engineering
University of North Carolina at Charlotte
Charlotte, USA
Email: b.chowdhury@uncc.edu

*Abstract—* **To mitigate the uncertainty of variable renewable resources, two off-the-shelf machine learning tools are deployed to forecast the solar power output of a solar photovoltaic system. The support vector machines generate the forecasts and the random forest acts as an ensemble learning method to combine the forecasts. The common ensemble technique in wind and solar power forecasting is the blending of meteorological data from several sources. In this study though, the present and the past solar power forecasts from several models, as well as the associated meteorological data, are incorporated into the random forest to combine and improve the accuracy of the day-ahead solar power forecasts. The performance of the combined model is evaluated over the entire year and compared with other combining techniques.**

## I. INTRODUCTION

The wind and solar energy resources have created operational challenges for the electric power grid due to the uncertainty involved in their output in the short term. The intermittency of these resources may adversely affect the operation of the electric grid when the penetration levels of these variable generations are high. Thus, wherever the variable generation resources are used, it becomes highly desirable to maintain higher than normal operating reserves and efficient energy storage systems to manage the power balance in the system. The operating reserves that use fossil fuel generating units should be kept to a minimum in order to get the maximum benefit from the deployment of the renewable resources. Therefore, the forecast of these variable generations becomes a vital tool in the operation of the power systems and electricity markets [1].

As in wind power forecasting, the solar power also consists of a variety of methods based on the time horizon being forecasted, the data available to the forecaster and the particular application of the forecast. The methods are broadly categorized according to the time horizon in which they generally show value. Methods that are common in solar power forecasting include Numerical Weather Prediction (NWP) and Model Output Statistics (MOS) to produce forecasts, as well as hybrid techniques that combine ensemble forecasts and Statistical Learning Methods [2]. Applying machine learning techniques directly to historical time series of solar photovoltaic (PV) production associated with NWP outcomes have placed among the top models in the recent global competition of energy forecasting, GEFCom2014 [3]. Just to name a few of the machine learning tools, the artificial neural networks (ANN) and support vector regression (SVR), gradient boosting (GB), random forest (RF), etc. are believed to be the most common.

Hybrid models of two or more statistical and physical techniques are often combined to capture complex interactions and provide useful insights and better forecasts. In ref. [4], the authors implement a hybrid model that consists of ARMA and ANN to forecast the solar irradiance by NWP data for 5 locations of a Mediterranean climate. They found the proposed model outperforms the naïve persistence model and improvement with respect to its core techniques as well. The study reported in ref. [5] presents the benefits of combining the data of solar irradiance that is derived from a satellite with ground-measured data to improve the intraday forecasts in the range up to six hours ahead. In ref [6], the authors combine satellite images with ANN outcomes to forecast the solar irradiance of leading time up to two hours for two sites in California.

In ref. [7], several statistical combining methods are used to combine multiple linear regression models for load forecasting, and the authors conclude that the regression combining technique is the best. While ref. [8] uses several statistical models to forecast the hourly PV electricity production for the next day at some power plants in France, the random forest (RF) has shown a superior performance. Ref. [9] also found the random forest has the best performance among others to predict the daily solar irradiance variability of four sites with different climatic conditions in Australia. The authors of [10] used random forest with other models as well to forecast the solar power in GEF2014 competition. They indicate that the random forest and the gradient boosting technique are the most accurate models. In the aforementioned studies, RF is not implemented there as a combining method; it is a standalone forecasting model that depends on its own trees.

The commonly used ensemble technique in wind and solar power forecasting is to blend the weather data from several sources. As in ref. [11], the authors compare several data-

**Note:** This is a pre-print of the full paper that published in *Innovative Smart Grid Technologies, North America Conference*, 2017, which can be referenced as below:
 M. Abuella and B. Chowdhury, "Random Forest Ensemble of Support Vector Regression for Solar Power Forecasting," in *Proceedings of Innovative Smart Grid Technologies, North America Conference*, 2017.

driven models using input data from two NWPs and building two artificial hybrid and stochastic ensemble models based on ANN, the model that combines multiple models outperforms the rest of the models. It points out that the ensemble is enhanced by including forecasts with similar accuracy, but generated from NWP data of higher variance and different data-driven techniques. Ref. [12] uses Ensemble Prediction System (EPS) to produce weather scenarios by running multiple initials to quantify the uncertainty and then produce the probabilistic solar power forecasts of sites in Italy. Ref. [13] applies a physical post processing and ANN to improve the solar irradiance forecasts of one and two days ahead.

The majority of the existing research literature on combining forecasting methods of solar forecasts do not include the previously generated forecasts to boost the model performance. It can be useful to add these past models' outcomes as well into the ensemble learning methods. The research team from the National Renewable Energy Laboratory (NREL) and IBM Thomas Watson Research Center [17] deployed and tested several machine-learning techniques to blend three NWPs outcomes. They conclude that the ensemble approaches that take into account the diversity and the state parameters of the models provide lower errors in the solar irradiance forecasting. Although these studies forecast the solar irradiance at different sites in the U.S., the time period is limited since they do not investigate the performance of the different seasons over the entire year.

Using RF as a combining method of other models in solar power is scarce. This paper adopts RF to combine the forecasts for a PV system and investigates the performance throughout a complete year. The rest of the paper is organized as follows: Section II describes the ensemble method that is used to build and combine the SVR models. Section III discusses the methodology of combining the solar power forecasts. Section IV presents the results and the evaluation of the model. Section V provides the conclusions.

## II. Ensemble Learning

The algorithms that use decision trees can be useful to combine the different models' outcomes efficiently. This ensemble approach combines all the outputs from variant models besides the features, such as the weather data that allow the ensemble method to find the associative rules to determine the best output. For instance, if the weather is sunny, then model A performs best, and its output should be selected; otherwise model B should be selected, and so on. This ensemble learning approach has shown very promising results in numerous machine learning benchmarks. For more details on this topic, i.e., ensemble learning and its techniques as bagging, boosting, stacking, and Bayesian averaging, the interested reader may refer to [16], The general diagram of combining the models is shown in Fig.1.

### A. Random Forest

Since the classification and regression trees (CART) use the bagging principle of the ensemble learning, and they are built by the same data, these trees are sometimes correlated and statistically dependent. Consequently, to make the trees more variant and uncorrelated, Breiman [17] proposed that each split of the bagged tree should be grown by a random number of features and observation samples. Hence, this method is called the random forest (RF).

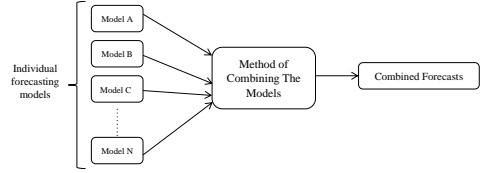

Fig.1. General diagram of combining different models

Three parameters are required to be set in RF, the number of trees $B$ (forest size), $m$ the number of predictors out of $p$ available variables (features) that are randomly chosen to be used for each split, and the minimum number $n_{min}$ of observations per node (leaf size).

The random forest building algorithm [16] has three major steps.

- Create $B$ sample datasets of size $N$ from the training data; these sample datasets can be replaced and overlapped.

- For each sample dataset, grow a random forest tree $T_b$, by repeating the following steps for each terminal node, until the minimum node size $n_{min}$ is reached:

   I. Select $m$ predictors at random from the $p$ variables.

   II. Pick the best predictor among the $m$ selected predictors for the split-point.

   III. Split this point (node) into two daughter nodes by setting certain decision rules.

- Finally, find the ensemble of the trees $\{T_b\}_1^B$, where $B$ is the number of trees in the random forest.

The prediction of a given point $x$ of the response variable can be obtained by averaging the individual tree's outputs:

$$\hat{f}_{RF} = \frac{1}{B}\sum_{b=1}^{B} T_b(x) \qquad (1)$$

The ensemble learning algorithm repeatedly assembles the input data to create regression trees that best fit the relationship between the features and the output. This process of decorrelation of the trees makes the random forest outcomes less variable, and hence more reliable.

## III. Modeling

### A. Data Description

The solar power system is in Australia and has a latitude 35°16'30"S, longitude 149°06'49"E, altitude 595m. The panel type is Solarfun SF160-24-1M195, consisting of 8 panels, its nominal power of 1560W, and panel orientation 38° clockwise from the north, with panel tilt of 36°. The weather forecasts data and the available measured solar power data starts from April 2012 to May 2014, as shown in Fig.2.

**Note:** This is a pre-print of the full paper that published in *Innovative Smart Grid Technologies, North America Conference*, 2017, which can be referenced as below:
M. Abuella and B. Chowdhury, "Random Forest Ensemble of Support Vector Regression for Solar Power Forecasting," in *Proceedings of Innovative Smart Grid Technologies, North America Conference*, 2017.

Fig.2. (a) The weather data and their numeral codes,
(b) available data size

| No. Vars. | Variable Name |
|---|---|
| 1 | Cloud Water Content |
| 2 | Cloud Ice Content |
| 3 | Surface Pressure |
| 4 | Relative Humidity |
| 5 | Cloud Cover |
| 6 | 10m- U Wind |
| 7 | 10m- V Wind |
| 8 | 2-m Temperature |
| 9 | Surface solar radiation down |
| 10 | Surface thermal radiation down |
| 11 | Top net solar radiation |
| 12 | Total precipitation |
| 13 | Heat Index |
| 14 | Wind speed |

(a)

| No. | Month | Year |
|---|---|---|
| 1 | April | 2012 |
| 2 | May | 2012 |
| 3 | June | 2012 |
| 4 | July | 2012 |
| 5 | August | 2012 |
| 6 | September | 2012 |
| 7 | October | 2012 |
| 8 | November | 2012 |
| 9 | December | 2012 |
| 10 | January | 2013 |
| 11 | February | 2013 |
| 12 | March | 2013 |
| 13 | April | 2013 |
| 14 | May | 2013 |
| 15 | June | 2013 |
| 16 | July | 2013 |
| 17 | August | 2013 |
| 18 | September | 2013 |
| 19 | October | 2013 |
| 20 | November | 2013 |
| 21 | December | 2013 |
| 22 | January | 2014 |
| 23 | February | 2014 |
| 24 | March | 2014 |
| 25 | April | 2014 |
| 26 | May | 2014 |

(b)

Normalizing the data is of paramount importance since the scale used for the values for each variable might be different. The best practice is to normalize the data and transform all the values to a common scale, as shown in equation (2).

$$X_{Scaled} = a + \frac{[x - \min(X)]}{[\max(X) - \min(X)]} * \{b - a\} \quad (2)$$

Where x is a sample from data variable X, {a, b} is the desired range of the normalized data, such as {0, 1}, and X (min, max)=the minimum and maximum values of the observed data.

There is also a standardization technique, especially when the variance of the data is high, which is making the data to have a zero mean and a unit standard deviation, as follows:

$$X_{standardized} = \frac{[x - \text{mean}(X)]}{\text{std}(X)} \quad (3)$$

*B. Model Building*

The different models are achieved by using different parameterization within the same model, i.e., support vector regression models (SVRs). Refer to [18] for more details about this model. The models are built as in Fig.3, where the available data is divided into two sets, one dataset consisting of all 26 months and another dataset consisting of the most recent 12 months only, as shown in Fig.2.a. Then, each of these datasets is used to build 12 SVR models based on two normalization techniques as in Equations (2) and (3), after that two different SVR's hyper-parameters of C and Gamma are also chosen, and different combination of weather variables are used to produce 12 models from each two datasets to get the total of 24 SVR models.

Fig.3. Construction of 12 SVR models from a dataset. Another 12 SVRs from the other dataset. The total number of SVR models is 24

The forecast methodology is shown in Fig.4.a. The forecasting day should be excluded from the data, while the rest of the available data is used to train the SVR models. The super learner is the random forest where the available weather and the previous forecasts are blended together to find the associative rules to achieve better solar power forecasts for the next day, as shown in Fig.4.b, and they should be as close as possible to the observed solar power that minimize the forecast errors of that day.

Fig.4. Demonstrates (a) the 24 hours ahead forecasting and (b) the combining methodology scheme for May 31st.

RF does not need cross-validation to estimate the parameters because it has a built-in estimation of accuracy. The parameter selection is carried out by the wrapping strategy or a greedy search for the best evaluation results among the available training data, the parameters search of RF are shown in Fig.5, [number of trees $B$=100, samples $m$=6 (i.e.,18/3), leaf size $n_{min}$ =5]. It is worth mentioning that a change with a reasonable range of these parameters does not affect the RF performance significantly. Thus, the robustness and flexibility of RF are the main advantages of this ensemble method.

Fig.5. The search for random forest parameters. (a) The forest size, (b) the features number at each node, (c) the leaf size (minimum number of samples at each node).

IV. RESULTS AND EVALUATION

The following metrics are used to evaluate the accuracy of the forecasts and the model performance: graphs, Root Mean Square Error (RMSE) as calculated by (4), and a comparison with other combining methods. For comparison purposes, the simple average combining method (Avg.), and the best model (Model 4) out of the 24 SVRs are adopted to evaluate the performance of the combined forecasts. Also, the improvement or the skill factor is used to compare the performance of the combined forecasts with respect to the other methods as in (5).

$$RMSE = \sqrt{\frac{1}{n}\sum_{i=1}^{n}(Y_i - \hat{Y}_i)^2} \quad (4)$$

where $\hat{Y}$ is the forecast of the solar power and $Y$ is the observed value of the solar power. $\hat{Y}$ and $Y$ are normalized to the nominal installed capacity of the solar power system, $n$ is the number of hours - it can be day hours or month hours. The objective is to minimize the RMSE for all forecasting hours to

**Note:** This is a pre-print of the full paper that published in *Innovative Smart Grid Technologies, North America Conference*, 2017, which can be referenced as below:
M. Abuella and B. Chowdhury, "Random Forest Ensemble of Support Vector Regression for Solar Power Forecasting," in *Proceedings of Innovative Smart Grid Technologies, North America Conference*, 2017.

yield more accurate forecasts. If the training and testing of the model are carried out for just the daylight hours while filtering out the night hours (which have zero solar power generation), the RMSE should also be determined for these daylight hours only without including the night hours.

$$\text{Improvement rate (\%)} = \frac{(Other\ method - Ensemble\ Method)}{Other\ Method} * 100 \quad (5)$$

The best model (model-4) is made of normalization technique (A), SVR's hyper-parameters C=10 gamma =8, and the original 14 weather inputs by using a dataset of all available months.

$$\text{Best Model (4)} = \text{All-months dataset + Normalize (A)} + \text{Parm10\_08 + Orig14ins} \quad (6)$$

In the graphical illustration, the month of October is chosen for comparison of all SVRs models with other combined models as shown in Fig.6. For the aggregation of daily RMSEs over the whole month, it is obvious the ensemble forecasts (black line) has a lower RMSE than SVR models and the simple average combining method (dark blue line). However, in a few days, the best SVR model (model-4) or the average method could be more accurate, as in 6th, 7th, 16th, 28th, and 29th days, when they have a lower RMSE.

The fluctuation in the daily RMSE can be seen among the 24 SVR models, and it's a normal trend in the forecasting models. However, the ensemble method produces more accurate combined forecasts and the performance of this combining model becomes more stable than in the individual SVR models.

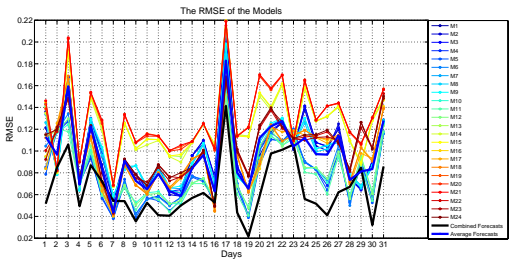

Fig.6. Daily RMSE of different models and combined forecasts, October

To get a broader evaluation of the combined forecasts performance, the comparison is conducted with the best model (model 4) and the simple average method over a complete year, as shown in TABLE I. It is clear that the ensemble method has lower monthly RMSEs. The improvement rate of the ensemble method over the other methods is calculated as in (5). In some months such as October, the ensemble method has an improvement rate of 18% and 28% over the best model and the average method respectively. For two months where the improvement rate is negative, such as in March the simple average and the best model are better than the ensemble, and also in June, the best model has the lowest monthly RMSE.

In general, the aggregated mean (i.e., the statistical average) of the monthly RMSEs indicates that the combined forecasts from the ensemble method have the most accurate forecasts (RMSE$_{mean}$=0.0725), while, the total improvement of the ensemble method over the average and the best model is 9% and 5 % respectively.

Notice that the monthly RMSE is calculated for all hours of the month where $n$ in (4) is not the day hour, but rather the month hours. Thus, it could be 744, or any other number of hours depending on the month.

TABLE I
RESULT OF DIFFERENT MODELS AND THE COMBINED FORECASTS

| Month | RMSE | | | Improvement of Ensemble Over: | |
|---|---|---|---|---|---|
| | Best Model (4) | Simple Average | Ensemble | Best Model (4) | Simple Avg. |
| June | 0.0734 | 0.0784 | 0.0764 | -4% | 3% |
| July | 0.0878 | 0.0877 | 0.0849 | 3% | 3% |
| August | 0.0859 | 0.0877 | 0.0833 | 3% | 5% |
| September | 0.0843 | 0.0921 | 0.0815 | 3% | 12% |
| October | 0.0851 | 0.0974 | 0.0701 | 18% | 28% |
| November | 0.0826 | 0.0923 | 0.0737 | 11% | 20% |
| December | 0.0747 | 0.0798 | 0.0643 | 14% | 19% |
| January | 0.0627 | 0.0670 | 0.0580 | 7% | 13% |
| February | 0.0772 | 0.0811 | 0.0730 | 5% | 10% |
| March | 0.0800 | 0.0822 | 0.0846 | -6% | -3% |
| April | 0.0645 | 0.0653 | 0.0640 | 1% | 2% |
| May | 0.0560 | 0.0559 | 0.0561 | 0% | 0% |
| **Aggregated Mean** | **0.0762** | **0.0806** | **0.0725** | **5%** | **9%** |

The Random Forest is a nonlinear model and a black-box and difficult to analyze. In order to get an idea about the performance of the model, a statistical analysis is conducted. Firstly, the importance estimation of the RF inputs (features) is found. This would be available since one of the algorithm steps is to estimate the best features to split the nodes, as explained in Section II.A. October and March are chosen for this analysis, because in these months, the largest change in the ensemble method performance occurs. The features are the weather variables and the 24 SVRs' outcomes. As shown in Fig.7 and Fig.8, the features do not have the similar pattern of importance; thereby, in the data training by RF, they would give different results and different performance.

Secondly, a statistical analysis by finding the standard deviation and the correlation between the SVR models' outcomes is conducted over the full year. Fig.9 presents the histograms of this statistical metrics.

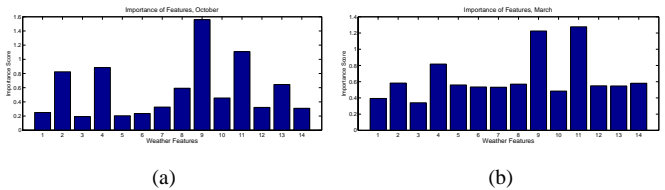

(a)           (b)

Fig.7. The estimation of weather features importance by Random Forest, (a) October, (b) March

**Note:** This is a pre-print of the full paper that published in *Innovative Smart Grid Technologies, North America Conference*, 2017, which can be referenced as below:
M. Abuella and B. Chowdhury, "Random Forest Ensemble of Support Vector Regression for Solar Power Forecasting," in *Proceedings of Innovative Smart Grid Technologies, North America Conference*, 2017.

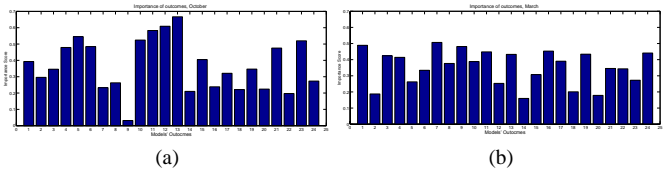

Fig.8. The estimation of models' outcomes importance by Random Forest, (a) October, (b) March

For October, the standard deviation is high, and hence the correlation of the SVRs' outcomes, and thus the ensemble method, has the best performance. However, in the last months - March, April, and May, the standard deviation is low and the correlation is high, and the performance of the ensemble method in these months is not as good as the other months, as indicated in TABLE I.

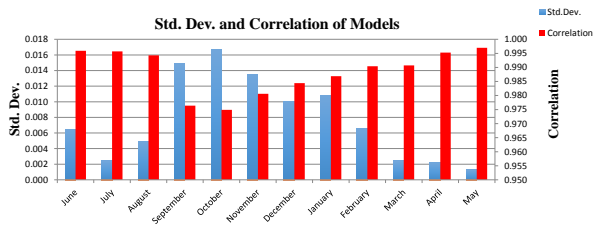

Fig.9. The standard deviation and the correlation of different models

## V. CONCLUSION

Combining the forecasts by the random forest leads to more accurate forecasts throughout the year. These combined forecasts are produced from an intelligent weighting approach that takes into account the weather situations, the past forecast of the forecasting models (24 SVRs), and their different temporal horizons - these all are used as associative rules in the ensemble method to yield accurate forecasts and a stable performance. The simple average as a combining method is not enough to get better forecasts since it does not capture or represent all the varieties in the weather data, and hence the solar power forecast. The forecasting errors are inherited mostly from the NWP model errors, and some of the errors resulting from the technical degrading of the physical systems (PV modules efficiency, orientation, etc.). Adding the past generated forecasts increases the accuracy of the combined forecasts.

**Mohamed Abuella** (S'14) received his Bachelor of Technology degree from College of industrial Technology, Misurata, Libya in 2008, and M.S. degree from Southern Illinois University Carbondale in 2012. He is currently a PhD student in the Department of Electrical and Computer Engineering at University of North Carolina at Charlotte. His research interest is in planning and operations of electrical power systems.

**Badrul Chowdhury** (S'83–M'87–SM'93) received the B.S. degree from the Bangladesh University of Engineering and Technology, Dhaka, Bangladesh, in 1981, and the M.S. and Ph.D. degrees from the Virginia Polytechnic Institute and State University, Blacksburg, VA, USA, in 1983 and 1987, respectively, all in electrical engineering. He is currently a Professor with the Department of Electrical and Computer Engineering with joint appointment with the Department of Systems Engineering and Engineering Management, University of North Carolina at Charlotte, Charlotte, NC, USA. His current research interests include power system modeling, analysis and control, and renewable and distributed energy resource modeling and integration in smart grids. Dr. Chowdhury is a Member of Tau Beta Pi and Phi Kappa Phi.

**Note:** This is a pre-print of the full paper that published in *Innovative Smart Grid Technologies, North America Conference*, 2017, which can be referenced as below:
M. Abuella and B. Chowdhury, "Random Forest Ensemble of Support Vector Regression for Solar Power Forecasting," in *Proceedings of Innovative Smart Grid Technologies, North America Conference*, 2017.